\documentclass[twoside]{article}

\usepackage{PRIMEarxiv}
\usepackage[utf8]{inputenc} % allow utf-8 input
\usepackage[T1]{fontenc}    % use 8-bit T1 fonts
\usepackage{hyperref}       % hyperlinks
\usepackage{url}            % simple URL typesetting
\usepackage{booktabs}       % professional-quality tables
\usepackage{amsfonts}       % blackboard math symbols
\usepackage{nicefrac}       % compact symbols for 1/2, etc.
\usepackage{microtype}      % microtypography
\usepackage{lipsum}
\usepackage{fancyhdr}       % header
\usepackage{graphicx}       % graphics
\graphicspath{{media/}}     % organize your images and other figures under media/ folder

\usepackage[numbers]{natbib}

\usepackage{multirow}
\usepackage{float}
\usepackage{subfig} 
\usepackage{amsmath}
\usepackage{mathtools}
\usepackage{bm}
\usepackage{multirow}
\usepackage{enumitem}
\usepackage[ruled,vlined,linesnumbered,noend]{algorithm2e}
\usepackage[table,xcdraw]{xcolor}

\usepackage{combelow}

\usepackage{xcolor}

\title{Advancements in eHealth Data Analytics through Natural Language Processing and Deep Learning}

\author{
  Elena-Simona Apostol and Ciprian-Octavian Truic{\u{a}} \\
  National University of Science and Technology Politehnica Bucharest, Bucharest, Romania \\
  \texttt{elena.apostol@upb.ro, ciprian.truica@upb.ro,}
}

\begin{document}
\maketitle

\begin{abstract}
The healthcare environment is commonly referred to as "information-rich" but also "knowledge poor". Healthcare systems collect huge amounts of data from various sources: lab reports, medical letters, logs of medical tools or programs, medical prescriptions, etc. These massive sets of data can provide great knowledge and information that can improve the medical services, and overall the healthcare domain, such as disease prediction by analyzing the patient's symptoms or disease prevention, by facilitating the discovery of behavioral factors for diseases. Unfortunately, only a relatively small volume of the textual eHealth data is processed and interpreted, an important factor being the difficulty in efficiently performing Big Data operations. In the medical field, detecting domain-specific multi-word terms is a crucial task as they can define an entire concept with a few words. A term can be defined as a linguistic structure or a concept, and it is composed of one or more words with a specific meaning to a domain. All the terms of a domain create its terminology. This chapter offers a critical study of the current, most performant solutions for analyzing unstructured (image and textual) eHealth data. This study also provides a comparison of the current Natural Language Processing and Deep Learning techniques in the eHealth context. Finally, we examine and discuss some of the current issues, and we define a set of research directions in this area.\end{abstract}


\maketitle

\section{Introduction}

The healthcare environment is commonly referred to as "information-rich" but also "knowledge poor". Healthcare systems collect huge amounts of data from various sources: lab reports, medical letters, logs of medical tools or programs, medical prescriptions, etc.
These massive sets of data can provide great knowledge and information which can improve the medical services, and overall the healthcare domain, such as disease prediction by analyzing the patient’s symptoms or disease prevention, by facilitating the discovery of behavioral factors for diseases.
Unfortunately, only a relatively small volume of the textual eHealth data is processed and interpreted, an important factor being the difficulty in efficiently performing Big Data operations.

This chapter offers a critical study of the current, most performant solutions for analyzing unstructured eHealth data.
From patient records, health plans, or insurance information to medical images, the majority of medical data generated is unstructured. Unstructured data is generally defined as data for which it is difficult or impossible to create a typical database schema. 

In the medical field, detecting domain-specific multi-word terms is a crucial task as they can define an entire concept with a few words. A term can be defined as a linguistic structure, a concept, and it is composed of one or more words with a specific meaning to a domain. All the terms of a domain create its terminology. Using as a baseline the terms in the terminology, the medical datasets are divided into linguistic terms and then are represented into vectors of identifiers.
This transformation is done using an algebraic model denoted as Vector Space Models. The current most performant vector space models are the word embeddings, that consider contextual features, determines domain-specific terms, and if necessary, also may correct text misspelling before outputting the final vector of identifiers. In this chapter, we provide a critical analysis of different vector space models, together with their applicability for medical data. 

The output of a vector space model is then fed to learning models. 
As tremendous amount of data is generated in the medical field, the need to apply Big Data Deep Learning techniques to make use of this information becomes more and more acute.
Therefore, in this chapter, we also provide a description and comparison of the current Deep Learning techniques, together with the set of medical-specific tasks that each presented Deep Learning model solves.

Finally, the deep learning models can also be applied to medical images with the aim: \textit{i)} of determining specific objects, such as tumours, or \textit{ii)} of establishing relationships between the features extracted from an image dataset and some corresponding classes, e.g. used in prediagnosis of different diseases. Subsequently, at the end of this chapter, we present a brief discussion of deep learning models for medical images.

\section{Natural Language Processing for eHealth data}
Text analytics consists of a set of techniques that apply software algorithms in order to understand the content of unstructured textual data, more precisely of written language. Some techniques, such as rule-based classification, are using a set of rules or logical statements defined by specialists. 
Natural language processing (NLP) employs more complex techniques that consider the context and the meaning of words, and not only searching for individual terms. As a result, such approaches have, in most cases, better metrics, i.e., accuracy. In eHealth, NLP solutions are very often used in the medical staff in tasks such as screening, diagnosing treatment, or patient monitoring. To this end, the NLP techniques are added to complex eHealth adaptive systems, i.e., computer-aided diagnostic systems and data-driven decision support systems. In the Big Data context, these systems are fed with large amounts of data collected from different sources, which they then use to learn how to provide certain services. 
An example is improving the efficiency in determining the diagnosis or the right treatment, by combining the current patient medical results with information from his/her historical healthcare data and the information learnt from other similar cases.

\begin{figure}[!ht]
    \centering
    \includegraphics[width=1\linewidth]{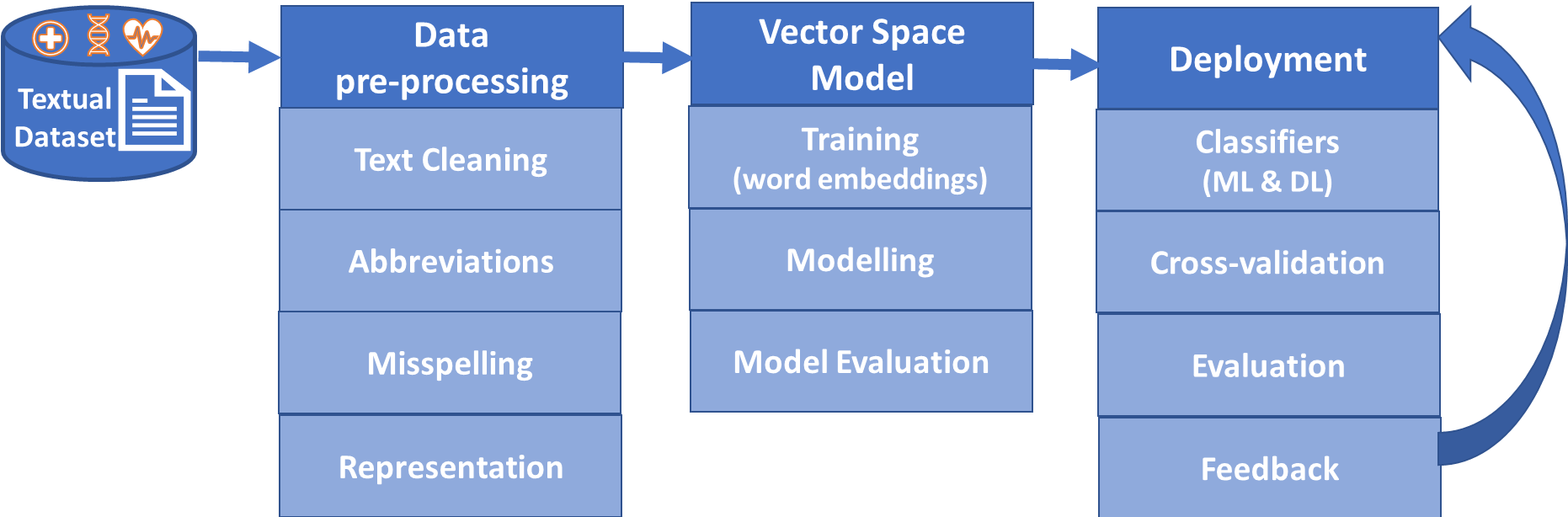}
    \caption{The Pipeline for textual data processing}
    \label{fig:pipeline}
\end{figure}

In order to obtain useful information and to model it according to the needs of the eHealth application or service, NLP utilises a processing pipeline.
Depending on the desired output, the NLP pipeline may consist of different specific steps, but they always group into three major steps: Preprocessing, Vector Space Model and Deployment (Figure \ref{fig:pipeline}).

During the preprocessing step, the text collection or the corpus is separated into linguistic units, e.g., words, expressions, sentences etc. From them, the most important information is extracted, information that will be then given to the language model to deduce new meanings from it. Different sub-steps can be added to the preprocessing stage. For example, when considering data gathered from physicians' observations and annotations, some words may be wrongly spelt, and thus a spell-checking sub-step can be introduced.

In the second step, the building of the space-vector or term-vector model is carried out. A space vector model represents the linguistic terms as vectors of weights. 
The result of this step will be fed to the language model in the Deployment step.

The final major step of the NLP pipeline is building the language model. The main objective of a language model is to estimate the probability distribution of the different linguistic units. There are two classes of language models: count-based and continuous-space. 

The count-based models, suchlike statistical language models, usually imply estimating the probability of the occurrence of particular word sequences based on samples of training datasets. There are several known disadvantages of this type of models. One important drawback is that most of the count-based models are not adaptive, as their results are based on exact matching of known patterns. As a result, finding new unknown words during testing is challenging. 

The continuous-space class refers to the neural language models. The idea behind these models is that the system learns an effective representation of the meanings of the words. Although this increases their complexity as opposed to classical count-based methods, it also gives them better performance. 

\subsection{State of the art in preprocessing eHealth text}
As mentioned earlier in this chapter, there is a large quantity of unstructured text in eHealth that is gathered from different sources, e.g., physicians' observations, annotations made to certain medical images, medical prescriptions, and so forth. Due to this heterogeneity, the collected medical data can not be directly fed to the learning model, and it must first go through a preprocessing step. Data preprocessing is the first step of the NLP pipeline, and on a broad view, it transforms text to vectors using a series of linguistic processing methods. In the end, it provides the input to the utilised vector space model. Although often the preprocessing step is overlooked, it can have a major influence on the accuracy of the vector space model and the learning model.

Depending on the available input corpus, there are different consistent and efficient variants of chaining the available text processing methods. However, a logical chaining for the preprocessing of the medical texts would be the following: \textit{i)} Text cleaning; \textit{ii)} Expansion of the abbreviations through abbreviation disambiguation methods; \textit{iii)} Misspelling correction; \textit{iv)} Text representation.

\subsubsection{ Abbreviation Disambiguation Methods}
Abbreviation Disambiguation is the task of identifying the intended long-form for an ambiguous short-form text, and it is part of the larger category of Word Sense Disambiguation (WSD) methods.
In recent years more and more abbreviations and acronyms have been introduced in the medical field, in order to facilitate the use of frequently referenced multi-word terms. With their increase in usage, several issues come to the fore. Firstly, one abbreviation or acronym can have multiple meanings, even in the same domain. Secondly, text understanding may be harder, as the abbreviations' definition is not always found alongside them.

As there is a large set of acronyms used in the medical field, the majority of the datasets containing acronyms are focused on this field, otherwise, the set is quite small or does not contain ambiguous acronyms. As a result of this, many proposed solutions for expanding abbreviations were considered for the medical field or at least used medical textual data in their experiments.

Considering the most popular studies for this subject, there are three major approaches: \textit{i)} algorithms based on a voting mechanism used in term recognition; \textit{ii)} unsupervised methods for recognition of multi-word terms; \textit{iii)} algorithms based on word embedding.

For the first approach, the voting system is used in order to select the most appropriate word sense from the multitude of outputs. This approach uses the weighted voting strategy based on rankings of a term produced by each term recognition algorithm. This algorithm can be implemented as part of the Automatic Term Recognition (ATR) or the Automatic Keyword Extraction (AKE) preprocessing steps. ATR is responsible for the extraction of technical terms from domain-specific language corpora, while AKE is the task of extracting the most relevant linguistic units. e.g. words, in a document with the purpose of automatic indexing~\cite{lossio2013combining}. 

Paper~\cite{zhang2008comparative} presents a comprehensive comparison of well known ATR and AKE methods that uses a voting system for abbreviation disambiguation. 
The experiments concluded that the voting system did not necessarily improve the performance of the algorithm due to the corpus’ quality and the specificity to multi-word terms. Another observation is that algorithms designed for multi-word terms will suffer from low precision when counting single-word terms or vice-versa.

FlexiTerm~\cite{spasic2018acronyms} is an example of an unsupervised method for recognition of multi-word terms that can be used for acronym recognition.
FlexiTerm performs term recognition in three steps: \textit{i)} Lexico-syntactic filtering - that is used in order to select multi-word candidates; \textit{ii)} Normalising the term candidates - in order to neutralize term variation; \textit{iii)} Computing the termhood. Termhood measures the association strength of a term to domain concepts. 
Likewise, since an acronym can be matched to multiple normalized term representatives, disambiguation needs to be performed. This process is accomplished by comparing potential normalised term representatives with respect to their frequency of occurrence and selecting the most frequent one as the most plausible full form. In the case of a tie, the comparison is made using the terms’ length measured by the number of tokens. In the unlikely event that an acronym remains ambiguous, brute-force is used to select the first term in alphabetical order.

A limitation of FlexiTerm is that it fails to classify some word formations that contain an acronym, e.g., ''exacerbation of chronic obstructive pulmonary disease'' and ''exacerbation of COPD'' should be considered equivalent, as COPD is not a standalone term.

The most efficient solutions for abbreviation disambiguation use Word Embedding techniques combined with back-off methods based on string similarity, e.g. \cite{li2015acronym}, \cite{charbonnier2018using}, \cite{wu2015clinical}. A word embedding is a learned representation of textual data that allows words with very similar connotations to have a similar representation.

Paper \cite{wu2015clinical} proposes a supervised machine learning-based word sense disambiguation method for clinical data. Their solution used a dataset consisting of unlabeled clinical notes, in order to train the neural word embeddings, and two manually annotated clinical datasets for testing. The training was performed using a Support Vector Machine (SVM)~\cite{cortes1995support} neural network. SVM is a supervised discriminative binary classifier that determines the best separation hyperplane to group the labeled points from a dataset. This classifier achieves state-of-the-art performance in many solutions for word sense disambiguation. This can also be observed for the results presented in paper~\cite{wu2015clinical}, where the average accuracy exceeds $0.92$ for all the datasets and regardless of the chosen set of features for the model.  

\subsubsection{Misspelling correction} \label{sec:misspelling}
In this section, we focus on a particular type of medical datasets, namely the data that come from the observations and annotations of physicians or different categories of medical reports. Since such textual documents are written in many cases under time pressure, the collected raw dataset may contain misspelt words which, if not corrected, will be considered as new terms by the NLP algorithms.

Automated detection and correction of misspellings is a popular research subject in many domains. Likewise, in recent years this topic has gained interest in the medical field, e.g. \cite{lai2015automated}, \cite{hussain2016identification}, \cite{dziadek2017improving}. The simplest techniques involve dictionary lookup, n-gram analysis and isolated-word error correction, whereas the most complex solutions are based on context-dependent misspellings correction.

n-gram analysis uses a probabilistic language model for predicting the next item of an n-gram. n-grams are consecutive sequences of items from a text corpus. These items can be letters, syllables or words. For automatic recognition of misspellings based on n-gram analysis, the basic idea is the usage of letter-frequency statistics to detect unusual sequences of characters, which can be indicators of possible errors. 

Systems based on isolated-word spelling correction consider that most of the spelling errors consist of tiny mistakes, such as the insertion or deletion of a letter, transposed or switched letters or the use of another letter in place of the correct one. Therefore, they calculate the minimum edit distance between words as the number of needed transformations necessary for a misspelt word to reach its correct variant.

The state-of-the-art solutions for context-dependent misspellings correction are using word embedding models to learn the association between misspellings and their corresponding correct words. Such solutions, based on word embedding, were also designed for the medical field, e.g., \cite{yazdani2019automated}. It is worth noting that the majority of current word embedding models, e.g., \cite{Pinter2017}, do not explicitly support misspellings correction, and they require as input data a set of pre-trained embeddings. One linguistic model that provides misspelling correction functionalities is MOE (Misspelling Oblivious Word Embeddings).
  
MOE (Misspelling Oblivious Word Embeddings)~\cite{piktus2019misspelling} is an NLP model for efficient learning of word representations and sentence classification. This model is extended with a supervised task that embeds misspellings close to their correct variants, called the spell correction loss. 
Practically, the spell correction loss is a typical logistic function which applies a scalar product between the sum of vectors with sub-words obtained from the misspelt and its corresponding correct word.

At the opposite pole regarding the supervised task offered by MOE, is the solution proposed in paper \cite{fivez2017unsupervised}. The solution uses an unsupervised context-sensitive spelling correction method.
This method generates candidates for the found misspellings and uses neural embeddings to rank them according to their semantic fit. The method was compared with two other tools, one of them being Hun-Spell, an open-source spell checker used by many browsers. The experimental results show that the proposed method greatly outperforms the baseline tools, although a more interesting comparison would be with another embeddings based solution, such as MOE.

\subsection{Efficient Vector Space Models for Medical Data}
The Vector Space model is an algebraic model that represents textual data as vectors of identifiers. The vector space dimension is its cardinality (i.e. number of vectors) over the base field. For textual data, the dimension may represent the terms from a dictionary. Let's consider a simple example, i.e. for building a vector space model, "A virus is a microorganism. A virus invades living cells.". After removing the stopwords like a, is, etc., the text keywords are extracted, e.g. virus, microorganism, etc. Using these keywords, one can build a simple vector space model based on the calculated weight of each keyword, e.g., the vector has a magnitude of 2 in the "virus" direction. Such an example can be considered part of traditional count-based textual models of which Bag of Words~\cite{Yogarajan2020} models belongs.
Although some solutions from this family of models are effective in extracting features from textual data (i.e., especially for small texts of the same domain), information such as structure, semantics and context are not considered. More sophisticated models provide these features as a vector representation of words, also known as embeddings. In the following subsections, we will present novel solutions to build the Vector Space Model using Word Embeddings.

\subsubsection{Training the Word Embedding Model}

The training of a word embeddings model is achieved by fitting the model to the specific textual corpus and by tuning its hyper-parameters.
For model fitting, it is necessary to find out how many iterations should be applied to obtain a sufficient performant word embedding model, without overfitting.
The model’s hyper-parameters are configuration dependent values that are setup empirically before training, and can be considered the following: \textit{i)} the input or output size; \textit{ii)} the number of epochs for the model; \textit{iii)} whether or not it is desired to have an early stopping (e.g., by monitoring the loss on the dataset); \textit{iv)} the number of artificial neurons from the network.
Training a good word embedding model is highly dependent on the chosen set of hyper-parameters.

The training of the word embeddings network can be done with a subset from the original corpus. However, there are many medical datasets, publicly available, that can be used for training the embeddings network, e.g., \textit{CHDS} - Child Health and Development Studies datasets are intended to research how disease and health pass down through generation\footnote{\url{http://www.chdstudies.org/research/information\_for\_researchers.php}}; \textit{Kent Ridge Biomedical Datasets} - High-dimensional datasets in the biomedical field \footnote{\url{http://leo.ugr.es/elvira/DBCRepository/}}.

Although it is advisable to train your own word embeddings with datasets from the utilised domain, i.e., medical field, this step is highly compute-intensive as it requires many processing resources. And likewise, it is highly time-consuming, e.g., the average time for one epoch can be around $4$-$5$ hours, of course, it depends a lot on the chosen network embedding.
Of course, there is the possibility of using pre-trained word embeddings. The pre-trained word embeddings are then loaded into the embedding network layer. 
Such pre-trained embeddings are offered for download on different modelling tools and many languages. For example, Google made available pre-trained models on over $100$ billion words from Google News corpus \footnote{\url{https://code.google.com/archive/p/word2vec}}.

A disadvantage of using these pre-trained embeddings is that they do not provide very accurate results for domain-specific tasks, as they are trained on generic datasets. To overcome this limitation, several solutions, that add term specific information to pre-trained word embeddings, were proposed.
In paper~\cite{patel-etal-2017-adapting} is proposed such a method that adds information from medical coding datasets to different pre-trained word embeddings. Through rigorous testing, the authors showed that the modified word embeddings give an improvement in f-score by $1\%$ on their private medical dataset.

\subsubsection{Modelling. Feature Extraction}

As mentioned in the previous subsection, word embeddings networks provide representations of words in a dense vector space, while also considering contextual and semantic information. Consequently, word embedding models are considered state of the art in textual feature extraction. Following, we will present some of the most popular models and tools for feature extraction using word embeddings.

\subsubsection*{Word2Vec Embedding}
Word2Vec \cite{mikolov2013efficient} is a two-layer neural network developed by a group of researchers from Google that builds a vectorized representation for the words from a very large corpus, based on the contexts in which they appear in that corpus.
The algorithm returns a numeric vectorial representation for each word, this representation having the property that words with similar context are closer to each other in the multi-dimensional vector space that is created. Starting from this property, one could compute the semantic similarity of two concepts by evaluating the cosine distance between the vectorial representation of these concepts.

Word2Vec utilizes two types of model architectures: CBOW (Continuous Bag-Of-Words) model and Skip-gram model. The CBOW model accounts for the textual dataset vocabulary, representing documents as a set of pairs of continuous word-multiplicity. 
In the Skip-gram model, documents are represented as sequences of words with gaps between them. The input to the neural network is the target word and the output layer is replicated multiple times to accommodate the chosen number of context words. Thus, this model manages to embed in the word representation its linguistic context.
The two models mirror each other while both preserve context. The CBOW model uses multiple neighbouring words to preserve the context for a target word, while the Skip-gram model uses a word to preserve the context for multiple targeted neighbouring words.

Word2Vec is still one of the most used network embeddings in eHealth data analytics, due to its simplicity in use and the fact that it is very intuitive. Also, an important factor is that it requires little memory, as data need light pre-processing and can be directly streamed into the model. 
For example, in paper~\cite{he2019classifying} is presented an efficient solution for identifying relations in clinical records. This task is not trivial, as clinical records may contain more than two medical concepts, most of the time comprised of several words. Their solution uses Word2Vec for building the embeddings that are then fed to a proposed Deep Neural network.

\subsubsection*{FastText Embedding} \label{sec:fasttext}
FastText~\cite{joulin2016fasttext} is built upon the Word2Vec model and improves it by providing high-quality embeddings by employing methods that are also time-efficient. Since its release, FastText has constantly increased in popularity, recently also being used for medical corpora.

In paper~\cite{soares2019medical} the authors utilise the FastText model to develop high-quality Spanish embeddings for the biomedical domain. These embeddings were made generally available by the authors. The training of their in-domain embeddings was done using two corpora: \textit{i)} a large corpus (i.e., SciELO database, which contains full-text articles primarily in English, Spanish and Portuguese); \textit{ii)} a smaller and focused corpus (i.e., Wikipedia Health, consisting of four main categories: Pharmacology, Pharmacy, Medicine and Biology). During hyper-parameters tuning they set the size of the word vector to $300$ and the number of epochs to $20$. The model was then used to identify pharmacological substances, compounds and proteins in clinical texts.

The FastText skipgram model is also used in paper~\cite{fivez2017unsupervised}, presented in Section~\ref{sec:misspelling} that deals with misspellings. In this case, FastText is used for creating vector representations of the candidates for the correct variants of the misspellings that are absent from the trained embedding space.

\subsubsection*{BERT Embedding}
Bidirectional Encoder Representations from Transformers (BERT)~\cite{DBLP:conf/naacl/DevlinCLT19} is a word embedding model based on a multilayer bi-directional deep network.
It uses a transformer deep neural network with up to $24$ layers, as opposed to the usual two-layer network used in other solutions, i.e., Word2Vec, FastText. 
Transformers networks use attention mechanisms that collect information about the relevant context of a given linguistic term, and then encode that context in the vector representation for that term.

A limitation of the BERT model is the relatively small size of the textual input block. Because of this, a relatively long clinical record will be split into multiple parts, and extrinsic tasks, as prediction, are done separately for each part. Once the prediction, or another extrinsic task, is applied for all the textual parts, the final probability results from their aggregation.

There are several variants of BERT, trained and fine-tuned for domain-specific text, this also being the case for the medical field, i.e., BioBERT and ClinicalBERT.

BioBERT~\cite{lee2020biobert} utilises almost the same architecture as BERT, the main alterations being made in the pre-training and fine-tuning steps. BioBERT is additionally pre-trained with two large biomedical corpora, namely the PubMed abstracts and the PMC full-text articles. Likewise, this model is fine-tuned for three fundamental biomedical specific tasks: \textit{i)} Named entity recognition (NER) which deals with recognizing the domain-specific terms in the medical corpora; \textit{ii)} Relation extraction (RE) which provides classifying relations of medical named entities; and \textit{iii)} Question answering (QA) which deals with answering NLP questions, given related text. For each of these specific tasks, BioBERT outperforms the current state of the art embedding models, including the original BERT model.

ClinicalBERT~\cite{DBLP:journals/corr/abs-1904-05342} was developed concurrently with BioBERT. This solution is built on top of the BERT neural network. As BioBERT, it is pre-trained with textual data from the medical domain, although the ClinicalBERT corpus is composed of large sets of clinical notes, and not of biomedical articles. ClinicalBERT is fine-tuned for mining clinical notes for readmission prediction. A comprehensive comparison of the current clinical BERT embedding, including BioBERT and ClinicalBERT, is carried out in article~\cite{Alsentzer2019}.

\subsubsection*{Sense-Disambiguation Embedding}
sense2vec~\cite{DBLP:journals/corr/TraskML15} is a new embeddings model specifically designed to disambiguate between word senses. Bassically, sense2vec provides multiple embeddings for a word based on the sense of that word. As opposed to traditional word embeddings, it analysis the context of a word and tries to assign more adequate vector for that word.

In paper~\cite{RiveraZavala2018} is used a combination of both word embeddings and sense-disambiguation embeddings. The features of both embedding models are concatenated and fed to a learning model, resulting in accurate detection of discontinuous entities, as well as of overlapping or nested entities. The learning model is used for knowledge recognition in eHealth documents.The proposed solution is performant, even without handcrafted features or specific domain knowledge. 

\subsubsection{Model Evaluation}

Word embeddings are evaluated using extrinsic methods that require apriori knowledge about the task that is solved, e.g., in Natural Language Processing the embeddings are evaluated on specific tasks where the correct outcome is already known such as part-of-speech tagging and name entity recognition. When working with medical data, word embedding evaluation must also be done using intrinsic methods, as, in some cases, there is no predefined target outcome. 

In the medical field, there are several standard datasets used in intrinsic evaluation, such as semantic similarity and relatedness, i.e., the UMLS similarity and relatedness datasets. UMLS(Unified Medical Language System)\footnote{\url{https://www.nlm.nih.gov/research/umls/index.html}} consists of manually annotated terms for similarity and relatedness.   

However, the state of the art intrinsic evaluation for word embeddings does not require apriori knowledge, such as the UMLS datasets. It focuses on studying the variation between two models of the target word's neighbors~\cite{Pierrejean2018}. The nearest neighbors of a target word is determined using cosine similarity. The variation between two models is computed using the degree of nearest neighbor variation for each model. The steps for computing the variance for the nearest neighbors for two words embedding models $M_1$ and $M_2$ are:
\begin{itemize}
\item extract the common vocabulary $V = V_1 \cup V_2$, where $V_1$ and $V_2$ are the  vocabulary for $M_1$, respectively $M_2$
\item determine the $k$ nearest neighbors ($knn$) of each term $w$ for both models, i.e., ${knn}_{M_{1}}(w)$ and ${knn}_{M_{2}}(w)$
\item compute the variation $var_{M_1, M_2}$ (Equaton~\eqref{eq:var_sim}) for the common neighbors among the $k$ nearest neighbors.
\end{itemize}

\begin{equation}\label{eq:var_sim}
{var}_{M_{1}, M_{2}}^{k}(w) = 1 - \frac{ || {knn}_{M_{1}}(w) \cap {knn}_{M_{2}}(w)|| }{k}
\end{equation}

The variation is a score in the interval $[0, 1]$. Values closer to the upper bound denote that the two models are drastically different. 

The parameter $k$ must be carefully chosen as it impacts the selection of the nearest neighbor of a term. Setting a small value for $k$ can result in omitting relevant neighbors while setting a large value can result in extracting loosely relevant neighbors. Experimental validation proved that the best value for $k$ is $25$, but we recommend performing hyper-parameter tuning for determining the best $k$ when training on custom datasets, i.e., medical records.

Although the intrinsic evaluation is more important in the case of many medical corpora, the majority of solutions for medical data provides both intrinsic and extrinsic evaluations. For example, considering the example discussed in subsection \ref{sec:fasttext} that proposed high-quality Spanish embeddings for the biomedical domain ~\cite{soares2019medical}, the authors detailed both the intrinsic and extrinsic evaluation. For the intrinsic evaluation, the authors used a traditional approach that requires apriori knowledge, but unfortunately, there are no annotated Spanish datasets for the medical domain. Hence, they adapted a similarity dataset from English to Spanish. We believe that a more efficient solution would have been the use of the variation-based model~\cite{Pierrejean2018}. As for the extrinsic evaluation of the proposed embeddings network, the authors deployed an artificial neural network with the purpose of identifying different substances, compounds and proteins from an open-access medical dataset. Finally, the authors offer a comparison of the extrinsic evaluation between a general domain embeddings and the one proposed by them. The results show that their solution is a bit better than the generic one, considering all the performance metrics, i.e., Accuracy, Precision, Recall and $F_{1}$ score (for their formulas see subsection \ref{sec:classEval}).

\subsection{An Evaluation of Deep Learning Architectures in the Medical Field} \label{sec:DLClass}

The last element of the NLP Pipeline is the Deployment phase.
In the following subsections, we detail the state of the art methods and models corresponding to this phase, highlighting how they are applied for processing medical textual data.  

\subsubsection{Deep-Learning Classifiers} 
The large collected datasets of unstructured medical data play an important role in the diagnosis process and provide a comprehensive view of the treatment and patient health status, but at the same time, it makes the process of analyzing data more difficult. 

In this context of the massive ever-growing of eHealth data, the traditional Machine Learning methods prove to be in some degree inefficient and inaccurate. Deep Learning algorithms evolved from traditional Machine Learning neural networks, however, they are used to build more complex and large neural networks. Likewise, the state of the art Natural Language Processing solutions applied on large scale eHealth data are powered by such Deep Learning algorithms.

Some of the most popular deep neuro-network architectures Learning algorithms used in eHealth solutions are: \textit{i)} Convolutional Neural Network (CNN)~\cite{he2019classifying} \cite{arifoglu2019detection} \cite{bui2019incorporated},  \textit{ii)} Recurrent Neural Networks (RNN)~\cite{hosseini2019sraki} \cite{yang2020identifying}, \textit{ii)} Long Short-Term Memory Networks (LSTMs)~\cite{rajeev2019intelligent}
In the following subsections, we provide short descriptions of these neuro-network architectures along with discussions about how they are applied in the medical field.

\subsubsection*{Convolutional Neural Networks}
Convolutional neural networks (CNNs) are a class of neural networks that are mainly used in visual imagery. It can take an input image and assign importance to aspects or objects in the image so as to differentiate one from another. One of the differences between regular neural networks and CNNs is that CNN explicitly assumes that the input is an image, which makes the preprocessing required much lower than for other classification algorithms~\cite{krizhevsky2012imagenet}.

CNNs are inspired by the connectivity pattern of neurons in the visual cortex. 
A neuron responds to stimuli in a restricted region of the visual field. The fields overlap to cover the entire visual area.

A CNN consists of an input layer and an output layer, as well as multiple hidden layers: convolutional layer, pooling layer, and fully connected layer. The fully connected layer is exactly as the hidden layer in regular neural networks.
The convolution layer of the neural network is the core building block in CNNs that does most of the computation. A set of learnable filters forms the parameters for this layer.
Each filter can be seen as a window sliding across the input, and the only part of the input that is analyzed at a certain time is the one inside the window. The result of the analysis is, actually, the dot product between the entries in the filter and the input at the current position of the window.

The function of a pooling layer in a CNN is to progressively reduce the spatial size of the representation to lower the number of parameters and computation in the network, contributing also to controlling overfitting. That is why it is common to insert a pooling layer between consecutive convolution layers. The pooling layer operates independently on every depth slice of the input and resizes it spatially, using the MAX operation.

The pooling layer accepts a volume of size $W1$ x $H1$ x $D1$, where $W1$, $H1$, and $D1$ are the width, height, and depth of the input in the pooling layer. It also requires two hyperparameters $F$, the spatial extent, and $S$, the stride, to produce a volume of size $W2$ x $H2$ x $D2$, where: $W_{2} = \frac{W_{1}-F}{S} + 1$, $H_{2} = \frac{H_{1}-F}{S} + 1$, $D_{2}=D_{1}$.

An important note is that, in practice, it is common to have $F = 2$ and $S = 2$, or $F = 3$ and $S = 2$, which results in overlapping pooling.

In general, CNN is applicable to datasets containing images, although it can also be used for text classification, in which case the depth of the convolutional layer is $1$ or $2$.

In recent biomedical papers, CNN was quite often used in combination with other deep networks for resolving different medical problems, e.g., detecting abnormal human behaviour, question answering, and knowledge base completion. 

In paper~\cite{arifoglu2019detection}, a CNN-based deep architecture is applied on medical datasets with the main purpose of detecting anomalies related to dementia, and indirectly to recognise daily life activities for the monitored patients.
Three deep networks are deployed and tested. The most complex and, at the same time, the most performant DL network is a combination of two-layer CNN with an LSTM layer (described later in this section), followed by two dense layers. In a dense layer network, each neuron is connected with all the neurons from the previous layer. The authors compared the performance of their solution with other Deep-Learning methods, i.e., LSTM only, and with more traditional Machine-Learning techniques, i.e., Hidden Markov Models (HMMs), Conditional Random Fields (CRF). Their solution outperforms the other methods in terms of Precision and Accuracy with mode than $0.04$, and more than $0.05$ respectively.

Article~\cite{he2019classifying} proposes a CNN architecture with a multi-pooling operation for medical relation classification on clinical records. The relation classification problem can be used for medical applications such as a question answering system. The authors classify the medical concepts relations into positive ones, e.g., "Test reveals medical problem", "Treatment worsens medical problem", and negative ones, e.g., "No relation between treatment and problem". The proposed architecture uses no external features, such as semantic role labeller, sentiment category, manually labelled patterns. The interval features are extracted via multi-pooling operation. The paper contains at the end a category-wise performance comparison with other neural network models, i.e., CNN-based models using max-pooling, SVM (Support Vector Machine).

\subsubsection*{Recurrent Neural Networks}
Recurrent Neural Networks (RNNs) are a class of neural networks that solve the problem of persisting the information about previous events by having loops inside them. This makes them capable of processing time sequences and learning sequences, which provided good results in speech to text comprehension, machine translation, and language modeling.
Due to the loop used in RNNs, the information is passed in the network from step to step, which can be thought of, actually, as duplicates of the same network, each one passing a message to a successor.

Some examples of RNNs are as follows:
\begin{enumerate}
    \item one to one – this type is actually equivalent to a regular neural network with fixed-sized input and fixed-sized output
    \item one to many – fixed sized input and sequence output, used, for example, in image captioning
    \item many to one – sequence input and fixed sized output, used in sentiment analysis where a sentence is classified as expressing a certain sentiment
    \item many to many – sequence input and sequence output, used in machine translation
     \item synced many to many – synced sequence input and sequence output, used in video classification where each frame needs to be labeled
\end{enumerate}

In the medical field, RNN-based architectures were used to provide performant solutions for medical application tasks that consider sequence and time features. An example of such specific tasks is human activity recognition. This task deals with predicting what a monitored patient is doing based on a movement trace, and it can be used in risk prevention and intervention medical applications for improving the health care quality.

One of the major limitations of the RNN networks is that the gradient of its loss function decays with time, almost exponentially. This implies that RNN networks are not performant in resolving problems that necessitate learning long-term temporal dependencies. 

Many of the current solutions for analysing unstructured medical data are no longer using the traditional RNNs, rather their architecture combines RNNs with layers of LSTM. LSTM was specially designed to overcome the RNNs gradient decay limitation. In the following subsection, we will provide a brief presentation of the LSTM deep networks.

\subsubsection*{Long Short-Term Memory Networks}
Being an extension of the RNN, the LSTM model implementation is similar to it, however, the latter model makes it easier to remember past data in the network memory. LSTM uses back-propagation in the training process, and it is intended to be used for data consisting of long time sequences.

As in the case of RNNs, the gradient of long-term components grows exponentially faster than the one for short-term temporal components.
The solution offered by LSTM architectures, also known as clipping the gradient, is based on thresholding the gradients' values before executing a gradient descent step.

For consistency with the recurrent network, the LSTM hidden layer size is also set to $256$. A recurrent layer applies a multi-layer long short-term memory RNN to an input sequence. This means that for each element in the input sequence, each layer computes as shown in Equation~\ref{eq:lstm}.

\begin{equation}\label{eq:lstm}
\begin{split}
i_{t} & = \sigma (W_{ii}x_{t} + b_{ii} + W_{hi}h_{t-1} + b_{hi}) \\
f_{t} & = \sigma (W_{if}x_{t} + b_{if} + W_{hf}h_{t-1} + b_{hf}) \\
g_{t} & = \tanh (W_{ig}x_{t} + b_{ig} + W_{hg}h_{t-1} + b_{hg})  \\
o_{t} & = \sigma (W_{io}x_{t} + b_{io} + W_{ho}h_{t-1} + b_{ho}) \\
c_{t} & = f_{t}\ast c_{t-1} + i_{t}\ast g_{t}                    \\
h_{t} & = o_{t}\ast \tanh(c_{t})
\end{split}
\end{equation}

where $h_{t}$ is the hidden state at time $t$, $c_{t}$ is the cell state at time $t$, $x_{t}$ is the input at time $t$, $h_{t-1}$ is the hidden state of the layer at time $t - 1$ or the initial hidden state at time 0, and $i_{t}$, $f_{t}$, $g_{t}$, $o_{t}$ are the input, forget, cell and output gates. $\sigma$ is the sigmoid function and $\ast$ is the Hadamard product.

In a multilayer LSTM, the input $x_{t}^{(l)}$ of the l-th layer, with $l \geq  2$,  is the hidden state $h_{t}^{(l-1)}$ of the previous layer multiplied by dropout $\delta_{t}^{(l-1)}$, where each $\delta_{t}^{(l-1)}$ is a Bernoulli random variable which is 0 with probability dropout, a given hyperparameter.

There are many LSTM architectures commonly used in textual processing, besides the traditional Unidirectional LSTM, e.g., \textit{i)} Bidirectional LSTM (BiLSTM); \textit{ii)} Stack LSTM; \textit{iii)} Attention-based LSTM. The Bidirectional LSTM consists of LSTMs that have their hidden state output merged at the same time. Such an architecture enables the neural network to control, at each time step, both forward and backward information.
A Stack LSTM architecture extends on two dimensions, both on the input horizontal direction (such as BiLSTM), but also on the depth direction, forming a grid like structure.
There are also other LSTM architectures, recently proposed, such as Fully‑connected LSTM~\cite{ji2020fully}.

The majority of the current state of the art solutions for analysing textual medical data, use LSTM classifiers either to train their embeddings, or for specific extrinsic tasks, or both. Although, the greatest usage of these neural networks is in tasks related to concept extraction.
In the following paragraphs, we present several current medical solutions that have embedded in their Deep-Learning architecture LSTM layers.

The architecture proposed in article~\cite{RiveraZavala2018} has two Bidirectional Long Short-Term Memory (BiLSTM) layers and a layer based on Conditional Random Field (CRF). It was designed for the task of Named Entity Recognition (NER) in biomedical texts. Nowadays, this task is of great importance in the medical field, as it is very important to identify quality information that refers to specific entities of interest in large amounts of documents.
In this solution, the BiLSTM network is used for obtaining the word and sense-disambiguation embeddings. The network returns for each linguistic terms six probabilities corresponding for each label from the applied encoding format.
The CRF layer takes as input the output of the last BiLSTM layer and obtains the most probable sequence of predicted labels, resulting in an improvement of the predictions' accuracy. 

In article~\cite{ji2020fully}, a new deep learning architecture is proposed for efficiently extracting medical concepts, i.e. symptoms, laboratory tests, and treatments, from large unstructured corpora. The architecture consists of Fully-Connected LSTM, proposed in this article, and Conditional Random Fields(CRF) layers.
The Fully-Connected LSTM (FC-LSTM) neural network is similar in structure with a stack LSTM, with the difference that the connection between the FC-LSTM layers is tighter. Furthermore, each FC-LSTM layer acts as a BiLSTM, as the training uses both forward and backward information.
The authors compared FC-LSTM, on medical concept extraction tasks, both with other types of classification methods (e.g., different CRF models) and with other classes of LSTM (e.g., Stack LSTM, Attention-based LSTM). In all the experiments, the FC-LSTM achieved the highest performance metrics, i.e., $F_{1}$-score, Precision. This may be due to the fact that FC-LSTM has short connection paths and can obtain information on multiple words concurrently.

\subsubsection*{Boltzmann Machines}
A Boltzmann Machine is a stochastic recurrent neural network that has two types of symmetrically fully-connected nodes: input or visible nodes and hidden nodes. The connections between nodes are weighted and can be represented as an undirected graph.  Unlike other artificial neural networks, A Boltzmann Machine does not have an output layer which gives them the non-deterministic feature. Instead, each node generates states as part of the entire system using an energy-based model and helps make stochastic decisions by being turned on or off.  For a learning problem, a Boltzmann Machine uses a simple learning algorithm that discovers interesting features in datasets composed of binary vectors. Unfortunately, the learning process of a Boltzmann Machine is very slow for network architectures with many layers of feature detectors as there are many small updates on the weights. To solve the complexity problem and make efficient weight updates, restrictions can be placed on the intralayer connections for both types of nodes, thus resulting in the Restricted Boltzmann Machine networks.

A Restricted Boltzmann Machine is probabilistic, unsupervised, generative deep learning networks. As Boltzmann Machine, the input and hidden layer are still present and the two layers are fully-connected. The difference between the two networks is in the intralayer connections, i.e., the connections between the input nodes and the ones between the hidden nodes, that are removed for the Restricted Boltzmann Machine architectures. In the field of Natural Language Processing of medical data, Restricted Boltzmann Machines have been successfully used for predicting patient diagnoses from Electronic Health Records~\cite{Zhang2018}

A Deep Boltzmann Machine architecture is an unsupervised, probabilistic, generative model with entirely undirected connections between different layers containing symmetrically connected stochastic binary nodes. It can be obtained by stacking multiple hidden layers. As in the case of Restricted Boltzmann Machines, the hidden layers and the visible layer do not contain any intralayer connections. Only neighboring layers are connected in a Deep Boltzmann Machine architecture. The architecture can be fine-tuned using backpropagation. Deep Boltzmann Machine networks have been used together with ontologies to learn semantic representations of words~\cite{Wang2016}. This approach can be used in the medical domain as the representations correspond to concepts at various semantic levels in a domain ontology.

A Deep Belief Network contains multiple stacks of Restricted Boltzmann Machines. Like in Deep Boltzmann Machine and Restricted Boltzmann Machines, there are no intralayer connections between the nodes. All the layers are fully-connected to their neighboring layers. The top two layers of a  Deep Belief Network contain symmetric undirected connections between the nodes of neighboring layers, similar to a Deep Boltzmann Machine. These layers form the associative memory of the architecture. Unlike Deep Boltzmann Machine, the lower layers contain directed acyclic connections between the nodes of neighboring layers. These layers are used to convert associative memory to observed variables. The input layer can accept data that is either binary or real. 

Deep belief network architectures are used for named entity recognition from electronic medical records~\cite{Li2018}.  Another application of Deep belief network architectures is extracting and fusing features of structured data and unstructured text medical data to deal with the highly nonlinear relationship between multimodal data~\cite{Hao2019}.

\subsubsection{Cross-validation}

Cross-validation is a statistical method used for estimating how well a machine learning model generalizes on an independent dataset. The basic method is the Holdout method that requires splitting the dataset into the training and testing set. The training set contains two-thirds of the dataset, while the testing contains the remaining third. Using this configuration, the model is build using the training set and then validated using the testing set. 

A generalized version of the Holdout method is the $k$-fold cross-validation. Using this method, the dataset elements are shuffled randomly and split into $k$ disjoint subsets of the same size. For each subset, a model is build using $k-1$ subsets as the training set and the remaining subset as the testing set. For each $k$ iteration the model is evaluated and, after the last iteration, an average evaluation score is computed. Once assigned to a subset, a data point will not change subsets for the duration of the entire procedure.  
In the case the dataset is imbalanced, the distribution of classes among the data point must be preserved to build a generalized machine learning model. To this means, the stratified cross-validation method is used to split the dataset into disjoint subsets that maintain class distribution. 

\subsubsection{Classifier Evaluation} \label{sec:classEval}

Machine learning classification is grouped into four categories~\cite{Sokolova2009}: \textit{i)} binary, \textit{ii)} multi-class, \textit{iii)} multi-labeled, and \textit{iv)} hierarchical. Depending on the type of classification, we need to define different measures to accurately evaluate a model.  

Binary classification assigns data point to one of two non-overlapping classes, i.e., $C_1$ and $C_2$, usually positive and negative. To evaluate the model we will need to build the confusion matrix where we will define the following types:
\begin{itemize}
    \item $TP$ (True Positive) is the number of observations that belong to the $C_1$ class and are classified correctly;
    \item $FN$ (False Negative) is the number of observations that belong to the $C_1$ class and are classified as $C_2$;
    \item $FP$ (False Positive) is the number of observations that belong to the $C_2$ class and are classified as $C_1$;
    \item $TN$ (True Negative) is the number of observations that belong to the $C_2$ class and are classified correctly.
\end{itemize}

Using the confusion matrix, for binary classification we can define the following evaluation metrics:
\begin{itemize}
    \item Accuracy ($A = \frac{TP + TN}{TP + FP + TN + FN}$) to measure the overall effectiveness of a classifier;
    \item Error Rate ($E = 1 - A $) to measure the proportion of incorrectly classified observations;
    \item Precision ($P = \frac{TP}{TP + FP}$) to measure the class agreement of the data labels within the positive labels given by a classifier;
    \item Recall or sensitivity ($R = \frac{TP}{TP + FN}$) to measure the effectiveness of a classifier to identify positive labels;
    \item Specificity ($S = \frac{TN}{TN + FP}$) to measure the effectiveness of a classifier to identify negative labels.
    \item F-Score ($F_{\beta} = (\beta^{2} + 1) \cdot \frac{P \cdot R}{\beta^2 \cdot P + R}$) to measure the relations between positive classes in the dataset and those given by a classifier.
\end{itemize}

Multi-class classification assigns a data point to one and only one class $C_i$, with $i=\overline{1, l}$. The $l$ classes are not overlapping. As in binary classification, an extended confusion matrix that holds values for each class $C_i$ is defined, i.e., will contain a separate $TP_i$, $FN_i$, $FP_i$, and $TN_i$ for each class. The quality of the model will be assessed by the same evaluation methods, i.e., Precision (Equation~\eqref{eq:pmc}), Recall (Equation~\eqref{eq:rmc}), and F-Score (Equation~\eqref{eq:fmc}), but in two ways: \textit{i)} macro-averaging ($M$) to measure the average per-class agreement of a class with those detected by the classifier, and \textit{ii)} micro-averaging ($\mu$) to measure the per-class agreement of a class with those detected by the classifier.

\begin{equation}\label{eq:pmc}
    P_{M} = \frac{1}{n}\sum_{i=1}{l}\frac{{TP}_i}{{TP}_i + {FP}_i} \,\,\,\, and \,\,\,\,
    P_{\mu} =\frac{\sum_{i=1}{l}{TP}_i}{\sum_{i=1}{l}({TP}_i + {FP}_i)}
\end{equation}

\begin{equation}\label{eq:rmc}
    R_{M} = \frac{1}{n}\sum_{i=1}{l}\frac{{TP}_i}{{TP}_i + {FN}_i} \,\,\,\, and \,\,\,\,
    R_{\mu} = \frac{\sum_{i=1}{l}{TP}_i}{\sum_{i=1}{l}({TP}_i + {FN}_i)}
\end{equation}

\begin{equation}\label{eq:fmc}
    F_{M} = (\beta^{2} + 1) \cdot \frac{P_{M} \cdot R_{M}}{\beta^2 \cdot P_{M} + R_{M}} \,\,\,\, and \,\,\,\,
    F_{\mu} = (\beta^{2} + 1) \cdot \frac{P_{\mu} \cdot R_{\mu}}{\beta^2 \cdot P_{\mu} + R_{\mu}}
\end{equation}

Multi-labeled classification assigns a data point to one and or more classes $C_i$, with $i=\overline{1, l}$. As in the case of multi-class classification, the classes are not overlapping. An observation $x_j$ belongs to class $C_i$ if $L_j[i] = 1$, where $L_j=\{L_j[i] | i=\overline{1,l}\}$ is a set of labels for the observation with $j=\overline{1,n}$ and $n$ the size of the dataset. The quality of the model is assessed through either partial or complete class label matching using an indicator function $I(L_{j}^{d}=L_{j}^{c})$, where $L_{j}^{d}$ represents the labels in the dataset for an observation $x_j$, and $L_{j}^{c}$ are the labels predicted by the classifier for the same observation. The Exact Match Ratio (Equation~\eqref{eq:emr}) measures the average per-label exact classification. The average per-label classification with partial matches can be measured using the Labelling F-Score (Equation~\eqref{eq:lfs}), while the average per-class classification with partial matches can be computed using the Retrieval F-Score (Equation~\eqref{eq:rfs}).

\begin{equation}\label{eq:emr}
    EMR = \frac{\sum_{j=1}^{n}I(L_{j}^{d}=L_{j}^{c})}{n}
\end{equation}

\begin{equation}\label{eq:lfs}
    F_{n} = \frac{\sum_{j=1}^{n}\frac{2\sum_{i=1}^{l}(L_{j}^{d} \cdot L_{j}^{c})}{\sum_{i=1}^{l}(L_{j}^{d} + L_{j}^{c})}}{n}
\end{equation}

\begin{equation}\label{eq:rfs}
    F_{l} = \frac{\sum_{i=1}^{l}\frac{2\sum_{j=1}^{n}(L_{j}^{d} \cdot L_{j}^{c})}{\sum_{j=1}^{n}(L_{j}^{d} + L_{j}^{c})}}{l}
\end{equation}

Hierarchical classification assigns a data point to one and only one class $C_i$, with $i=\overline{1, l}$. Each class can be divided into sub-classes ($C_{\downarrow}$) or grouped into super-classes $C_{\uparrow}$. The hierarchy is pre-defined and stable, i.e., does not suffer changes during classification. The quality measures evaluate descendant ($\downarrow$) or ancestor ($\uparrow$) performance by computing 
Precision (Equation~\eqref{eq:pdu}), 
Recall (Equation~\eqref{eq:rdu}), 
and F-Score (Equation~\eqref{eq:fdu}) on 
the correct data labels, i.e., $C_{\downarrow}^d$ and $C_{\uparrow}^d$, and the labels assigned by the classifier, i.e., $C_{\downarrow}^c$ and $C_{\uparrow}^c$.

\begin{equation}\label{eq:pdu}
    P_{\downarrow} = \frac{|C_{\downarrow}^d \cap C_{\downarrow}^c|}{|C_{\downarrow}^c|} \,\,\,\, and \,\,\,\,
    P_{\uparrow} = \frac{|C_{\uparrow}^d \cap C_{\uparrow}^c|}{|C_{\uparrow}^c|}
\end{equation}

\begin{equation}\label{eq:rdu}
    R_{\downarrow} = \frac{|C_{\downarrow}^d \cap C_{\downarrow}^c|}{|C_{\downarrow}^d|} \,\,\,\, and \,\,\,\,
    R_{\uparrow} = \frac{|C_{\uparrow}^d \cap C_{\uparrow}^c|}{|C_{\uparrow}^d|}
\end{equation}

\begin{equation}\label{eq:fdu}
    F_{\downarrow} = (\beta^{2} + 1) \cdot \frac{P_{\downarrow} \cdot R_{\downarrow}}{\beta^2 \cdot P_{\downarrow} + R_{\downarrow}} \,\,\,\, and \,\,\,\,
    F_{\uparrow} = (\beta^{2} + 1) \cdot \frac{P_{\uparrow} \cdot R_{\uparrow}}{\beta^2 \cdot P_{\uparrow} + R_{\uparrow}}
\end{equation}

\subsubsection{Feedback}

A model can incorrectly classify observation in a dataset. The evaluation results should be interpreted by an expert in the field. In the context of the medical domain, it is required that the expert has inside knowledge and details about the observations. For example, in the case where the observations are collected from patients, there should be apriori knowledge about the health history, physical and psychological, and other test results for each individual being used to construct the training dataset of the model. Thus, sensitivity and specificity must present values close to $100\%$. A model with high sensitivity does not present many false-negative results and manages to identify correctly individuals using a well-defined screening process. A model with a high specificity does not present many false-positive results and manages to identify correctly only the individuals that have the condition for which the model is built.

After a model is trained and evaluated, it needs to pass the golden standard. In the case of medicine, this standard test is the model on which we can extrapolate a procedure, whether or not it increases the number of correct screening of patients. Usually, an expert in the domain must validate the results and, if required, must propose the workflow and implementation of the procedure. If the procedure directly involves the treatment of patients, each patient recovery status must be under close observation to medical experts to intervene and adjust the recovery process. 

If the experts observe that the model does not perform accurately, they will give adjust the model through feedback and retrain the model. The feedback can contain but is not limited to new observations, new attributes that were not taken into account when the model was trained, a better resampling of the dataset, etc.

\section{Deep Learning techniques for medical images} \label{DL-image} 
Medical images, e.g., photographs, x-rays, Magnetic Resonance Imaging (MRIs) or Computer Axial Tomography (CAT) scans, can provide additional information for a decision system and they have become a core data type in the eHealth field due to their increasing availability in digital form. Nevertheless, they make the problem of understanding data even more complex as their raw content is not explicit for machine processing and processing them is a high computationally task. Also, taking into account the medical context, temporary constraints in obtaining the results can be imposed on the processing tasks. Likewise, in many cases, there are privacy issues regarding the manipulation of the medical image datasets.

For extracting useful information from medical images, different deep-learning models can be applied, as in the case of the textual processing. A more detailed description of these models is presented in Section~\ref{sec:DLClass}.
Some other techniques used in textual processing can also be applied to medical images, even if this field is in the early stages. For example, tagging techniques would be achieved by tagging all images of a certain patient with the same identifier. An obstacle would be represented by the fact that viewing and analyzing a medical image is constrained to using only the vendor equipment. But lastly, several standards have been created to allow the implementation of vendor-independent image viewers ~\cite{marconi2014big}.
Image searching techniques allow searching images by visual content using image search patterns or a language description of that pattern, a process known as computer-based image retrieval (CBIR).

All the image processing tasks can be classified in one of two major categories: segmentation and classification.
For both of the two classes, specific deep learning models can be applied. Segmentation tasks are used to identify different elements from an image, e.g. identify organs or lesions, which can then be applied in complex analysis.
In classification tasks, the model is trained to establish a functional relationship between the features extracted from the image dataset and the corresponding classes, defined in the model. Examples of classification tasks include \textit{i)}the detection of different types of malignant and benign tumours from medical images, \textit{ii)} the detection of malaria from blood smear slide images~\cite{gopakumar2019deep}. 

However, deep neural architectures can be used for other tasks than classification and segmentation, such as image pre-processing. An example of such a task is Image Denoising. A solution for efficient image denoising using a deep architecture is proposed in article~\cite{rajeev2019intelligent}. The deep architecture employs LSTM based Batch Normalization and RNN techniques. The input dataset consists of CT images of the lung. The proposed solution removes different types of noises, i.e., white, salt and pepper noises, without requiring apriori knowledge. In comparison with other deep learning architectures, i.e., based on CNN and Gaussian denoising CNN (DnCNN), the proposed solution obtained the highest accuracy.

The majority of deep learning solutions proposed for processing medical images use different combinations of CNNs and LSTMs layers. Although the CNN based architectures produce very good results in processing 2D medical images, e.g., \cite{frid2018gan}, they alone do not have state of the art performance in processing 3D medical images, as they are hard to optimize for 3D volumetric feature-based classification. Better performance can be obtained by adding LSTM layers to the architecture, e.g., in article~\cite{shahzadi2018cnn} a CNN-LSTM deep architecture is used for classifying, with high accuracy, 3D brain tumour in MR (Magnetic Resonance) images.

In recent years, Boltzmann Machine based networks began to be widely used in medical imaging, with state of the art performance. Restricted Boltzmann Machines are used intensively for segmentation tasks, as automatic identification of Barrett's esophagus from endoscopic images of the lower esophagus~\cite{Passos2019}. Deep Boltzmann Machine networks are used mostly in classification tasks, where they obtain extraordinary results. In article~\cite{Jeyaraj2019}, a Deep Boltzmann Machine network is used for classifying medical images to detect pre-cancerous and post-cancerous regions. The resulting performance metrics are very high in comparison with other solutions, i.e., as high as $95.5\%$ accuracy and $93.5\%$ sensitivity.

\section{Conclusions}

Large organizations like hospitals collect huge amounts of data about their patients every year, in various ways, e.g., lab reports, medical prescriptions, general files of the patient when entering the hospital, x-rays, Computer Axial Tomographies. Such massive sets of data can provide great knowledge and information which can improve the medical services, and overall the healthcare domain, such as disease prediction by analyzing the patient’s symptoms or disease prevention, by facilitating the discovery of behavioral factors which can turn into risk factors for disease (e.g., diet, alcohol consumption, environmental pollution or physical activities). 

Unfortunately, only a relatively small volume of medical data is processed and interpreted, an important factor being the difficulty in efficiently performing Big Data operations. Likewise, another important factor is that, often, real and up to date medical datasets, even if anonymous, are hard to obtain, due to privacy issues. 

In eHealth processing, the complexity of data analysis also arises from its heterogeneity, starting with the collection of individual data elements and moving to the fusion of multiple datasets. Therefore, in many cases, it is necessary to combine different types and formats of data gathered from disparate heterogeneous sources.

Hence the full process is a difficult and intensely computational one, often needing Cloud like processing power. 
Nevertheless, the possible outcomes have the ability to reveal new ways of preventing, detecting and treating diseases.
Fortunately, new and improved methods for knowledge extraction, from medical text and image, are continuously proposed.

% \setcitestyle{numbers} % set the citation style to ``numbers''.
\bibliographystyle{plainnat}  
\bibliography{main}

\begin{thebibliography}{45}
\providecommand{\natexlab}[1]{#1}
\providecommand{\url}[1]{\texttt{#1}}
\expandafter\ifx\csname urlstyle\endcsname\relax
  \providecommand{\doi}[1]{doi: #1}\else
  \providecommand{\doi}{doi: \begingroup \urlstyle{rm}\Url}\fi

\bibitem[Alsentzer et~al.(2019)Alsentzer, Murphy, Boag, Weng, Jindi, Naumann,
  and McDermott]{Alsentzer2019}
Emily Alsentzer, John Murphy, William Boag, Wei-Hung Weng, Di~Jindi, Tristan
  Naumann, and Matthew McDermott.
\newblock Publicly available clinical {BERT} embeddings.
\newblock In \emph{Proceedings of the 2nd Clinical Natural Language Processing
  Workshop}, pages 72--78, Minneapolis, Minnesota, USA, June 2019. Association
  for Computational Linguistics.
\newblock \doi{10.18653/v1/W19-1909}.
\newblock URL \url{https://aclanthology.org/W19-1909}.

\bibitem[Arifoglu and Bouchachia(2019)]{arifoglu2019detection}
Damla Arifoglu and Abdelhamid Bouchachia.
\newblock Detection of abnormal behaviour for dementia sufferers using
  convolutional neural networks.
\newblock \emph{Artificial intelligence in medicine}, 94:\penalty0 88--95,
  2019.

\bibitem[Bui et~al.(2019)Bui, Lee, and Shin]{bui2019incorporated}
Toan~Duc Bui, Jae-Joon Lee, and Jitae Shin.
\newblock Incorporated region detection and classification using deep
  convolutional networks for bone age assessment.
\newblock \emph{Artificial intelligence in medicine}, 97:\penalty0 1--8, 2019.

\bibitem[Charbonnier and Wartena(2018)]{charbonnier2018using}
Jean Charbonnier and Christian Wartena.
\newblock Using word embeddings for unsupervised acronym disambiguation.
\newblock In \emph{International Conference on Computational Linguistics},
  pages 2610--2619. Association for Computational Linguistics, 2018.

\bibitem[Cortes and Vapnik(1995)]{cortes1995support}
Corinna Cortes and Vladimir Vapnik.
\newblock Support-vector networks.
\newblock \emph{Machine learning}, 20\penalty0 (3):\penalty0 273--297, 1995.

\bibitem[Devlin et~al.(2019)Devlin, Chang, Lee, and
  Toutanova]{DBLP:conf/naacl/DevlinCLT19}
Jacob Devlin, Ming{-}Wei Chang, Kenton Lee, and Kristina Toutanova.
\newblock {BERT:} pre-training of deep bidirectional transformers for language
  understanding.
\newblock In Jill Burstein, Christy Doran, and Thamar Solorio, editors,
  \emph{Proceedings of the 2019 Conference of the North American Chapter of the
  Association for Computational Linguistics: Human Language Technologies,
  {NAACL-HLT} 2019, Minneapolis, MN, USA, June 2-7, 2019, Volume 1 (Long and
  Short Papers)}, pages 4171--4186. Association for Computational Linguistics,
  2019.
\newblock \doi{10.18653/v1/n19-1423}.

\bibitem[Dziadek et~al.(2017)Dziadek, Henriksson, and
  Duneld]{dziadek2017improving}
Juliusz Dziadek, Aron Henriksson, and Martin Duneld.
\newblock Improving terminology mapping in clinical text with context-sensitive
  spelling correction.
\newblock \emph{Informatics for Health: Connected Citizen-Led Wellness and
  Population Health}, 235:\penalty0 241--245, 2017.

\bibitem[Fivez et~al.(2017)Fivez, {\v{S}}uster, and
  Daelemans]{fivez2017unsupervised}
Pieter Fivez, Simon {\v{S}}uster, and Walter Daelemans.
\newblock Unsupervised context-sensitive spelling correction of clinical
  free-text with word and character n-gram embeddings.
\newblock In \emph{{BioNLP} 2017}. Association for Computational Linguistics,
  2017.
\newblock \doi{10.18653/v1/W17-2317}.

\bibitem[Frid-Adar et~al.(2018)Frid-Adar, Diamant, Klang, Amitai, Goldberger,
  and Greenspan]{frid2018gan}
Maayan Frid-Adar, Idit Diamant, Eyal Klang, Michal Amitai, Jacob Goldberger,
  and Hayit Greenspan.
\newblock Gan-based synthetic medical image augmentation for increased cnn
  performance in liver lesion classification.
\newblock \emph{Neurocomputing}, 321:\penalty0 321--331, 2018.

\bibitem[Gopakumar and Subrahmanyam(2019)]{gopakumar2019deep}
G~Gopakumar and Gorthi RK~Sai Subrahmanyam.
\newblock Deep learning applications to cytopathology: A study on the detection
  of malaria and on the classification of leukaemia cell-lines.
\newblock In \emph{Handbook of Deep Learning Applications}, pages 219--257.
  Springer, 2019.

\bibitem[Hao et~al.(2019)Hao, Usama, Yang, Hossain, and Ghoneim]{Hao2019}
Yixue Hao, Mohd Usama, Jun Yang, M.~Shamim Hossain, and Ahmed Ghoneim.
\newblock Recurrent convolutional neural network based multimodal disease risk
  prediction.
\newblock \emph{Future Generation Computer Systems}, 92:\penalty0 76--83, mar
  2019.
\newblock \doi{10.1016/j.future.2018.09.031}.

\bibitem[He et~al.(2019)He, Guan, and Dai]{he2019classifying}
Bin He, Yi~Guan, and Rui Dai.
\newblock Classifying medical relations in clinical text via convolutional
  neural networks.
\newblock \emph{Artificial intelligence in medicine}, 93:\penalty0 43--49,
  2019.

\bibitem[Hosseini et~al.(2019)Hosseini, Zhang, Uǧurbil, Moeller, and
  Ak{\c{c}}akaya]{hosseini2019sraki}
Seyed Amir~Hossein Hosseini, Chi Zhang, K{\^a}mil Uǧurbil, Steen Moeller, and
  Mehmet Ak{\c{c}}akaya.
\newblock sraki-rnn: accelerated mri with scan-specific recurrent neural
  networks using densely connected blocks.
\newblock In \emph{Wavelets and Sparsity XVIII}, volume 11138, page 111381B.
  International Society for Optics and Photonics, 2019.

\bibitem[Huang et~al.(2019)Huang, Altosaar, and
  Ranganath]{DBLP:journals/corr/abs-1904-05342}
Kexin Huang, Jaan Altosaar, and Rajesh Ranganath.
\newblock Clinicalbert: Modeling clinical notes and predicting hospital
  readmission.
\newblock \emph{CoRR}, abs/1904.05342, 2019.
\newblock URL \url{http://arxiv.org/abs/1904.05342}.

\bibitem[Hussain and Qamar(2016)]{hussain2016identification}
Faiza Hussain and Usman Qamar.
\newblock Identification and correction of misspelled drugs names in electronic
  medical records (emr).
\newblock In \emph{International Conference on Enterprise Information Systems},
  pages 333--338, 2016.

\bibitem[Jeyaraj and Nadar(2019)]{Jeyaraj2019}
Pandia~Rajan Jeyaraj and Edward Rajan~Samuel Nadar.
\newblock Deep boltzmann machine algorithm for accurate medical image analysis
  for classification of cancerous region.
\newblock \emph{Cognitive Computation and Systems}, 1\penalty0 (3):\penalty0
  85--90, sep 2019.
\newblock \doi{10.1049/ccs.2019.0004}.

\bibitem[Ji et~al.(2020)Ji, Chen, and Jiang]{ji2020fully}
Jie Ji, Bairui Chen, and Hongcheng Jiang.
\newblock Fully-connected lstm--crf on medical concept extraction.
\newblock \emph{International Journal of Machine Learning and Cybernetics},
  pages 1--9, 2020.

\bibitem[Joulin et~al.(2016)Joulin, Grave, Bojanowski, Douze, J{\'e}gou, and
  Mikolov]{joulin2016fasttext}
Armand Joulin, Edouard Grave, Piotr Bojanowski, Matthijs Douze, H{\'e}rve
  J{\'e}gou, and Tomas Mikolov.
\newblock Fasttext. zip: Compressing text classification models.
\newblock \emph{arXiv preprint arXiv:1612.03651}, 2016.

\bibitem[Krizhevsky et~al.(2012)Krizhevsky, Sutskever, and
  Hinton]{krizhevsky2012imagenet}
Alex Krizhevsky, Ilya Sutskever, and Geoffrey~E Hinton.
\newblock Imagenet classification with deep convolutional neural networks.
\newblock In \emph{Advances in neural information processing systems}, pages
  1097--1105, 2012.

\bibitem[Lai et~al.(2015)Lai, Topaz, Goss, and Zhou]{lai2015automated}
Kenneth~H. Lai, Maxim Topaz, Foster~R. Goss, and Li~Zhou.
\newblock Automated misspelling detection and correction in clinical free-text
  records.
\newblock \emph{Journal of Biomedical Informatics}, 55:\penalty0 188--195, jun
  2015.
\newblock \doi{10.1016/j.jbi.2015.04.008}.

\bibitem[Lee et~al.(2020)Lee, Yoon, Kim, Kim, Kim, So, and
  Kang]{lee2020biobert}
Jinhyuk Lee, Wonjin Yoon, Sungdong Kim, Donghyeon Kim, Sunkyu Kim, Chan~Ho So,
  and Jaewoo Kang.
\newblock Biobert: a pre-trained biomedical language representation model for
  biomedical text mining.
\newblock \emph{Bioinformatics}, 36\penalty0 (4):\penalty0 1234--1240, 2020.

\bibitem[Li et~al.(2015)Li, Ji, and Yan]{li2015acronym}
Chao Li, Lei Ji, and Jun Yan.
\newblock Acronym disambiguation using word embedding.
\newblock In \emph{AAAI Conference on Artificial Intelligence}, pages
  4178--4179, 2015.

\bibitem[Li et~al.(2018)Li, Shi, Gao, Wei, Zhu, Lin, Jiang, and Gao]{Li2018}
Wusuo Li, Shenghui Shi, Ziqiao Gao, Wei Wei, Qunxiong Zhu, Xiaoyong Lin,
  Daguang Jiang, and Shang Gao.
\newblock Improved deep belief network model and its application in named
  entity recognition of chinese electronic medical records.
\newblock In \emph{International Conference on Big Data Analysis}. {IEEE}, mar
  2018.
\newblock \doi{10.1109/ICBDA.2018.8367707}.

\bibitem[Lossio-Ventura et~al.(2013)Lossio-Ventura, Jonquet, Roche, and
  Teisseire]{lossio2013combining}
Juan~Antonio Lossio-Ventura, Clement Jonquet, Mathieu Roche, and Maguelonne
  Teisseire.
\newblock Combining c-value and keyword extraction methods for biomedical terms
  extraction.
\newblock In \emph{International Symposium on Languages in Biology and
  Medicine}, pages 45--49, 2013.

\bibitem[Marconi and Lehmann(2014)]{marconi2014big}
Katherine Marconi and Harold Lehmann.
\newblock \emph{Big data and health analytics}.
\newblock Crc Press, 2014.

\bibitem[Mikolov et~al.(2013)Mikolov, Chen, Corrado, and
  Dean]{mikolov2013efficient}
Tom{\'{a}}s Mikolov, Kai Chen, Greg Corrado, and Jeffrey Dean.
\newblock Efficient estimation of word representations in vector space.
\newblock In Yoshua Bengio and Yann LeCun, editors, \emph{1st International
  Conference on Learning Representations, {ICLR} 2013, Scottsdale, Arizona,
  USA, May 2-4, 2013, Workshop Track Proceedings}, 2013.

\bibitem[Passos et~al.(2019)Passos, de~Souza~Jr., Mendel, Ebigbo, Probst,
  Messmann, Palm, and Papa]{Passos2019}
Leandro~A. Passos, Luis~A. de~Souza~Jr., Robert Mendel, Alanna Ebigbo, Andreas
  Probst, Helmut Messmann, Christoph Palm, and Jo{\~{a}}o~Paulo Papa.
\newblock Barrett's esophagus analysis using infinity restricted boltzmann
  machines.
\newblock \emph{Journal of Visual Communication and Image Representation},
  59:\penalty0 475--485, feb 2019.
\newblock \doi{10.1016/j.jvcir.2019.01.043}.

\bibitem[Patel et~al.(2017)Patel, Patel, Golakiya, Bhattacharyya, and
  Birari]{patel-etal-2017-adapting}
Kevin Patel, Divya Patel, Mansi Golakiya, Pushpak Bhattacharyya, and Nilesh
  Birari.
\newblock Adapting pre-trained word embeddings for use in medical coding.
\newblock In \emph{{BioNLP} 2017}, pages 302--306. Association for
  Computational Linguistics, 2017.
\newblock \doi{10.18653/v1/W17-2338}.

\bibitem[Pierrejean and Tanguy(2018)]{Pierrejean2018}
Benedicte Pierrejean and Ludovic Tanguy.
\newblock Towards qualitative word embeddings evaluation: Measuring neighbors
  variation.
\newblock In \emph{Proceedings of the 2018 Conference of the North American
  Chapter of the Association for Computational Linguistics: Student Research
  Workshop}. Association for Computational Linguistics, 2018.
\newblock \doi{10.18653/v1/N18-4005}.

\bibitem[Piktus et~al.(2019)Piktus, Edizel, Bojanowski, Grave, Ferreira, and
  Silvestri]{piktus2019misspelling}
Aleksandra Piktus, Necati~Bora Edizel, Piotr Bojanowski, {\'E}douard Grave, Rui
  Ferreira, and Fabrizio Silvestri.
\newblock Misspelling oblivious word embeddings.
\newblock In \emph{Proceedings of the 2019 Conference of the North American
  Chapter of the Association for Computational Linguistics: Human Language
  Technologies, Volume 1 (Long and Short Papers)}, pages 3226--3234, 2019.
\newblock \doi{10.18653/v1/N19-1326}.

\bibitem[Pinter et~al.(2017)Pinter, Guthrie, and Eisenstein]{Pinter2017}
Yuval Pinter, Robert Guthrie, and Jacob Eisenstein.
\newblock Mimicking word embeddings using subword {RNNs}.
\newblock In \emph{Conference on Empirical Methods in Natural Language
  Processing}, pages 102--112. Association for Computational Linguistics, 2017.
\newblock \doi{10.18653/v1/D17-1010}.

\bibitem[Rajeev et~al.(2019)Rajeev, Samath, and
  Karthikeyan]{rajeev2019intelligent}
R~Rajeev, J~Abdul Samath, and NK~Karthikeyan.
\newblock An intelligent recurrent neural network with long short-term memory
  (lstm) based batch normalization for medical image denoising.
\newblock \emph{Journal of medical systems}, 43\penalty0 (8):\penalty0 234,
  2019.

\bibitem[{Rivera Zavala} et~al.(2018){Rivera Zavala}, Mart{\'{i}}nez, and
  Segura-Bedmar]{RiveraZavala2018}
Renzo~M. {Rivera Zavala}, Paloma Mart{\'{i}}nez, and Isabel Segura-Bedmar.
\newblock {A Hybrid Bi-LSTM-CRF model for knowledge recognition from ehealth
  documents}.
\newblock \emph{CEUR Workshop Proceedings}, 2172:\penalty0 65--70, 2018.
\newblock ISSN 16130073.

\bibitem[Shahzadi et~al.(2018)Shahzadi, Tang, Meriadeau, and
  Quyyum]{shahzadi2018cnn}
Iram Shahzadi, Tong~Boon Tang, Fabrice Meriadeau, and Abdul Quyyum.
\newblock Cnn-lstm: Cascaded framework for brain tumour classification.
\newblock In \emph{2018 IEEE-EMBS Conference on Biomedical Engineering and
  Sciences (IECBES)}, pages 633--637. IEEE, 2018.

\bibitem[Soares et~al.(2019)Soares, Villegas, Gonzalez-Agirre, Krallinger, and
  Armengol-Estap{\'e}]{soares2019medical}
Felipe Soares, Marta Villegas, Aitor Gonzalez-Agirre, Martin Krallinger, and
  Jordi Armengol-Estap{\'e}.
\newblock Medical word embeddings for spanish: Development and evaluation.
\newblock In \emph{Clinical Natural Language Processing Workshop}, pages
  124--133, 2019.

\bibitem[Sokolova and Lapalme(2009)]{Sokolova2009}
Marina Sokolova and Guy Lapalme.
\newblock A systematic analysis of performance measures for classification
  tasks.
\newblock \emph{Information Processing {\&} Management}, 45\penalty0
  (4):\penalty0 427--437, jul 2009.
\newblock \doi{10.1016/j.ipm.2009.03.002}.

\bibitem[Spasic(2018)]{spasic2018acronyms}
Irena Spasic.
\newblock Acronyms as an integral part of multi-word term recognition
  {\textendash} a token of appreciation.
\newblock \emph{{IEEE} Access}, 6:\penalty0 8351--8363, 2018.
\newblock \doi{10.1109/ACCESS.2018.2807122}.

\bibitem[Trask et~al.(2015)Trask, Michalak, and
  Liu]{DBLP:journals/corr/TraskML15}
Andrew Trask, Phil Michalak, and John Liu.
\newblock sense2vec - {A} fast and accurate method for word sense
  disambiguation in neural word embeddings.
\newblock \emph{CoRR}, abs/1511.06388, 2015.
\newblock URL \url{http://arxiv.org/abs/1511.06388}.

\bibitem[Wang et~al.(2016)Wang, Dou, and Lowd]{Wang2016}
Hao Wang, Dejing Dou, and Daniel Lowd.
\newblock Ontology-based deep restricted boltzmann machine.
\newblock In \emph{International Conference on Database and Expert Systems
  Applications}, pages 431--445. Springer International Publishing, 2016.
\newblock \doi{10.1007/978-3-319-44403-1_27}.

\bibitem[Wu et~al.(2015)Wu, Xu, Zhang, and Xu]{wu2015clinical}
Yonghui Wu, Jun Xu, Yaoyun Zhang, and Hua Xu.
\newblock Clinical abbreviation disambiguation using neural word embeddings.
\newblock In \emph{Proceedings of {BioNLP} 15}. Association for Computational
  Linguistics, 2015.
\newblock \doi{10.18653/v1/W15-3822}.

\bibitem[Yang et~al.(2020)Yang, Bian, Fang, Bjarnadottir, Hogan, and
  Wu]{yang2020identifying}
Xi~Yang, Jiang Bian, Ruogu Fang, Ragnhildur~I Bjarnadottir, William~R Hogan,
  and Yonghui Wu.
\newblock Identifying relations of medications with adverse drug events using
  recurrent convolutional neural networks and gradient boosting.
\newblock \emph{Journal of the American Medical Informatics Association},
  27\penalty0 (1):\penalty0 65--72, 2020.

\bibitem[Yazdani et~al.(2019)Yazdani, Ghazisaeedi, Ahmadinejad, Giti, Amjadi,
  and Nahvijou]{yazdani2019automated}
Azita Yazdani, Marjan Ghazisaeedi, Nasrin Ahmadinejad, Masoumeh Giti, Habibe
  Amjadi, and Azin Nahvijou.
\newblock Automated misspelling detection and correction in persian clinical
  text.
\newblock \emph{Journal of Digital Imaging}, pages 1--8, 2019.
\newblock \doi{10.1007/s10278-019-00296-y}.

\bibitem[Yogarajan et~al.(2020)Yogarajan, Gouk, Smith, Mayo, and
  Pfahringer]{Yogarajan2020}
Vithya Yogarajan, Henry Gouk, Tony Smith, Michael Mayo, and Bernhard
  Pfahringer.
\newblock Comparing high dimensional word embeddings trained on medical text to
  bag-of-words for predicting medical codes.
\newblock In \emph{Intelligent Information and Database Systems}, pages
  97--108. Springer International Publishing, 2020.
\newblock \doi{10.1007/978-3-030-41964-6_9}.

\bibitem[Zhang et~al.(2018)Zhang, Henao, Gan, Li, and Carin]{Zhang2018}
Yinyuan Zhang, Ricardo Henao, Zhe Gan, Yitong Li, and Lawrence Carin.
\newblock Multi-label learning from medical plain text with convolutional
  residual models.
\newblock In \emph{Machine Learning for Healthcare Conference}, volume~85 of
  \emph{Machine Learning Research}, pages 280--294, 2018.

\bibitem[Zhang et~al.(2008)Zhang, Iria, Brewster, and
  Ciravegna]{zhang2008comparative}
Ziqi Zhang, Jos{\'e} Iria, Christopher Brewster, and Fabio Ciravegna.
\newblock A comparative evaluation of term recognition algorithms.
\newblock In \emph{LREC}, volume~5, 2008.

\end{thebibliography}

\end{document}